\pdfoutput=1

\documentclass[11pt]{article}

\usepackage[]{ACL2023}

\usepackage{times}
\usepackage{latexsym}
\usepackage{array}
\usepackage{amssymb}
\usepackage{amsmath}
\usepackage{amsthm}
\usepackage{mathtools}
\usepackage{bbm}
\usepackage{bm}
\usepackage{multirow}
\usepackage{booktabs}
\usepackage{graphicx}
\usepackage{subfigure}
\usepackage{textcomp}
\usepackage{natbib}
\usepackage{enumitem}
\usepackage{pifont}
\usepackage{ulem}
\usepackage{cancel}
\usepackage{caption}

\usepackage[T1]{fontenc}

\usepackage[utf8]{inputenc}

\usepackage{microtype}

\usepackage{inconsolata}

\title{Interpreting Sentiment Composition with Latent Semantic Tree}
\author{
Zhongtao Jiang$^{1,2}$, Yuanzhe Zhang$^{1,2}$, Cao Liu$^3$, Jiansong Chen$^3$, Jun Zhao$^{1,2}$, Kang Liu$^{1,2}$\\
$^1$The Laboratory of Cognition and Decision Intelligence for Complex Systems,\\
Institute of Automation, Chinese Academy of Sciences\\
$^2$School of Artificial Intelligence, University of Chinese Academy of Sciences\\
$^3$Meituan\\
\texttt{\{zhongtao.jiang, yzzhang, jzhao, kliu\}@nlpr.ia.ac.cn}\\
\texttt{\{liucao, chenjiansong\}@meituan.com}
}

\begin{document}
\maketitle
\setlist{before=\normalfont,font=\normalfont\itshape, labelwidth =\widthof{\bfseries9}, leftmargin=!}
\setdescription{itemsep=0pt,partopsep=0pt,parsep=\parskip,topsep=5pt}
\begin{abstract}
As the key to sentiment analysis, sentiment composition considers the classification of a constituent via classifications of its contained sub-constituents and rules operated on them. Such compositionality has been widely studied previously in the form of hierarchical trees including untagged and sentiment ones, which are intrinsically suboptimal in our view. To address this, we propose semantic tree, a new tree form capable of interpreting the sentiment composition in a principled way. Semantic tree is a derivation of a context-free grammar (CFG) describing the specific composition rules on difference semantic roles, which is designed carefully following previous linguistic conclusions. However, semantic tree is a latent variable since there is no its annotation in regular datasets. Thus, in our method, it is marginalized out via inside algorithm and learned to optimize the classification performance. Quantitative and qualitative results demonstrate that our method not only achieves better or competitive results compared to baselines in the setting of regular and domain adaptation classification, and also generates plausible tree explanations\footnote{Data and code implementation is available at \url{https://github.com/changmenseng/semantic_tree}.}.
\end{abstract}

\section{Introduction}
Sentiment classification is a task to determine the sentiment polarity of a sentence \cite{yadav2020sentiment, dang2020sentiment}. Current researches on this task are gradually shifting from improving model performance to interpretability. As the most known stream, \textit{feature-based explanation} tries to figure out which input feature, say word, has the most influence on the prediction, in the form of the salience score or rationale, and in both self and post-hoc settings \cite{li2016understanding, ribeiro2016should, kim2020interpretation, lei2016rationalizing, bastings2019interpretable, de2020decisions}. However, this task requires sentiment composition \cite{polanyi2006contextual}, which is beyond the ability of these feature-based explanations.

\begin{figure}[t]
    \centering
    \subfigure[Untagged tree.]{
        \includegraphics[width=0.28\linewidth]{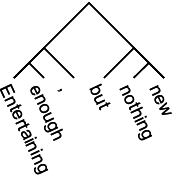}
        \label{fig:untagged_tree}
    }
    \subfigure[Sentiment tree.]{
        \includegraphics[width=0.28\linewidth]{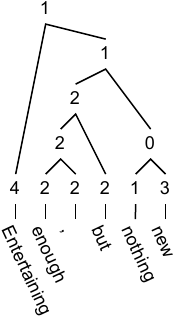}
        \label{fig:sentiment_tree}
    }
    \subfigure[Semantic tree.]{
        \includegraphics[width=0.28\linewidth]{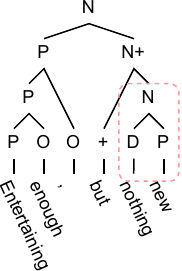}
        \label{fig:semantic_tree}
    }
    \caption{Different tree structures for explanining sentiment composition, where semantic tree can explain the sentiment composition in the inverted-V structure, as shown in the box of (c).}
    \label{fig:tree_structure}
\end{figure}

To be concrete, sentiment composition considers the classification of a constituent via 1) classifications of its contained sub-constituents and 2) rules operated on them \cite{moilanen2007sentiment}, as shown in Figure \ref{fig:semantic_tree}. Thus, the classification of a sentence is decomposed into hierarchical sentiment compositions of its sub-constituents. Such compositionality has been widely studied previously in the form of hierarchical trees including \textit{untagged tree} and \textit{sentiment tree}, as shown in Figure \ref{fig:tree_structure}. Untagged tree is usually modeled as a latent variable and learned via the task objective \cite{yogatama2016learning, maillard2018latent, choi2018learning, havrylov2019cooperative, chowdhury2021modeling}. Then, a TreeLSTM \cite{tai2015improved, zhu2015long} is adopted to encode the sentence following the hierarchy for the final prediction. However, untagged tree is limited because it can only explain the hierarchy but not give labels on all nodes. Sentiment tree takes a further step that every node within has a polarity score or label. As the most representative example, \citet{socher2013recursive} creates Stanford Sentiment Treebank (SST) that has sentiment tree annotation. Sentiment tree also appears as a post-hoc explanation giving hierarchical attribution scores \cite{chen2020generating, zhang2020interpreting}. However, in fact, not every constituent is sentimental, some of which are somewhat more functional. For example, while a negator ``not" is sentimentally neural, it can functionally flip the sentiment of a constituent. Sentiment labels are therefore not sufficient to explain such phenomenon.

To overcome those defects, we propose \textit{semantic tree}, a new tree form capable of explicitly and principally interpreting the sentiment composition. In the semantic tree, each node is assigned a label in \textit{semantic labels} including sentimental and functional ones, and each local inverted-V structure reveals the rule composing adjacent constituents, as shown in Figure \ref{fig:semantic_tree}. Inspired by \citet{dong2015statistical}, formally, the semantic tree is a derivation of a context-free grammar (CFG) \cite{chomsky1956three} defined by non-terminal symbols (semantic labels), terminal symbols (word vocabulary), rules, and root symbols (\textit{positive} and \textit{negative}). The challenge of designing such grammar lies in designing semantic labels and rules, which requires linguistic knowledge of sentiment composition. To address this, we follow previous work about sentiment composition \cite{polanyi2006contextual, moilanen2007sentiment, taboada2011lexicon} to carefully design 11 semantic labels and 62 rules. We believe the grammar could cover most cases in sentiment analysis, as shown in Table \ref{tab:composition_num}.

We aim to learn a model capable of extracting the semantic tree using data consisting of only sentence-label pairs, which is challenging because the semantic tree is latent without full annotation. To address this, we first build a semantic tree parser, and then marginalize out the semantic tree to induce a sentiment classifier to conduct supervised training on such data. Fortunately, this marginalization over the exponential tree space is computationally tractable resorting to the inside algorithm \cite{baker1979trainable}. This process could be abstracted as a module, namely sentiment composition module (SCM), which computes the compatibility of a prediction in the view of sentiment composition but not only pattern recognition. Accompanying an arbitrary neural text encoder with the proposed SCM, we can build a self-explanatory model that can not only predict the sentiment label but also generate a semantic tree as the explanation. To learn more plausible semantic trees, we further propose two extra objectives to guide the preterminals in the semantic tree, and to make the tree structure more syntactically meaningful.

We conduct experiments on three datasets including MR \cite{pang2005seeing}, SST2 \cite{socher2013recursive} and Amazon \cite{blitzer2007biographies} in the setting of regular and cross-domain classification. Quantitative and qualitative results demonstrate that our method not only achieves better or competitive results compared to baselines, and also generates plausible tree explanations.

\section{Method}

\subsection{Problem Formalization}
The dataset is a collection of tuples $\{(x^n,y^n)\}_{n=1}^N$, each of which contains a sentence $x\in\cal{V}^*$ and a sentiment label $y\in\cal{Y}$, where $\cal{V}$ is the word vocabulary and ${\cal Y}=\{P,N\}$ is the label set consisting of \textit{positive} ($P$) and \textit{negative} ($N$). The task goal is to learn a classifier $p(y|x)$. Since we hope to generate a semantic tree of the input sentence where the sentiment label is its root label, as shown in Figure \ref{fig:semantic_tree}, the objective classifier $p(y|x)$ is not directly parameterized by a discriminative model as usual. Instead, we define the classifier as the marginalization of a parser over the latent semantic tree, in which the parser could fulfill this purpose. Concretely, let ${\cal T}_x(y)$ be the set of all semantic trees rooted $y$. Naturally, we have:
\begin{equation}
p(y|x)=\sum_{t\in{\cal T}_x(y)}p(t|x)
\label{eq:formalization}
\end{equation}
where $p(t|x)$ is a semantic tree parser that accepts a sentence and generates a semantic tree. We can conduct supervised learning when the classifier $p(y|x)$ is obtained, where the parser $p(t|x)$ is implicitly learned in this process. After training, the model can do the prediction via the induced classifier $p(y|x)$, and generate the semantic tree to real the sentiment composition process of it.

The very first issue before solving the summation in Equation (\ref{eq:formalization}) is to formalize the semantic tree. For simplicity, we can assume that the label of a constituent is determined immediately by its sub-constituents, regardless of the surrounding context. Therefore, the semantic tree is viewed as a derivation of a CFG that defines specific semantic labels and composition rules. Now, two challenges remain: 1) How to properly define the CFG behind the semantic tree? 2) How to model the parser $p(t|x)$ and efficiently compute the classifier $p(y|x)$? We shall elaborate these two problems in Section 2.2 and Section 2.3, respectively.

\subsection{Sentiment Composition Grammar}
\label{sec:grammar}

The proposed semantic tree is described by a context-free grammar ${\cal G}$ consisting a quadruple including the non-terminal symbol set ${\cal N}$ (semantic label set), the terminal symbol set ${\cal V}$ (word vocabulary), the composition rule set ${\cal R}$ and the root symbol set ${\cal Y}$ ($P$ and $N$). While ${\cal V}$ and ${\cal Y}$ are obvious, the design of semantic labels (${\cal N}$) and composition rules (${\cal R}$) requires expert knowledge. Fortunately, previous works have concluded different types of compositions exhaustively \cite{polanyi2006contextual, moilanen2007sentiment, taboada2011lexicon}, inspiring us to design 11 semantic labels and 62 composition rules. We call the proposed grammar as a sentiment composition grammar (SCG).

\subsubsection*{Semantic Labels}
The defined 11 semantic labels include two types as follows:
\begin{description}
    \item[Sentimental labels] Including negative $N$, positive $P$, neutral $O$.
    \item[Functional labels] Including negator $D$, irrealis blocker $I$, priority riser $+$, priority reducer $-$, high negative $N^+$, high positive $P^+$, low negative $N^-$, low positive $P^-$.
\end{description}
We shall explain these labels together with composition rules later. 

\subsubsection*{Composition Rules} 
Formally, the composition rule is in the form of $\beta\rightarrow A$ ($A\in {\cal N}$, $\beta\in({\cal N}\cup {\cal V})^*$), which determines the label of a constituent given its sub-constituents\footnote{In the standard CFG, the rule is in the production form: $A\rightarrow\beta$. Since we want to model the sentiment composition, our rule is written in the equivalent converse way.}. We include three types of rules. 

The first one is binary rule in the form of $BC\rightarrow A$ ($A,B,C\in{\cal N}$). Binary rules are defined following common binary compositions, which mainly includes four types according to previous works and our observations. We now introduce each composition and its corresponding rules\footnote{Note that we assume binary rules to commutative, i.e., if we have rule $BC\rightarrow A$, then $CB\rightarrow A$ also holds. Thus, we only describe half of these rules in the following. Also, for compactness, we use symbol "/" to represent "or" operation. For example, $B\ C/D\rightarrow A$ means both $BC\rightarrow A$ and $BD\rightarrow A$ hold.}.

\begin{description}
\item[Polarity propagation] Propagating the polarity:
\begin{equation}
N\ O/N\rightarrow N,\ P\ O/P\rightarrow P,\ O\ O\rightarrow O
\end{equation}
\item[Negation] Flipping the non-neutral polarity ($P/N$) via a negator ($D$): 
\begin{equation}
D\ P\rightarrow N, D\ N\rightarrow P
\end{equation}
\item[Conflict Resolution] Resolving the conflict of non-neutral polarity constituents ($P/N$) by ranking their priorities based on priority modifiers ($+/-$). As a typical example, Figure \ref{fig:tree_structure} shows a contrastive conjunction \cite{socher2013recursive} structure, which the first and the second half of the sentence have opposite polarities. The connector ``but'' is a priority riser ($+$) that rises the priority of the second half sentence, which dominates the entire sentence priority. Similarly, there also exist priority reducer ($-$) such as ``although''. Thus, rules related to this composition includes those for priority modification: 
\begin{equation}
+\ P\rightarrow P^+, -\ P\rightarrow P^-
\end{equation}
and those for resolution: 
\begin{equation}
N\ P^+\rightarrow P, N^-\ P^+\rightarrow P
\end{equation}
We don't allow the polarity with priority ($N^+/N^-/P^+/P^-$) without a explicit modifier $+/-$, which a single word with non-neutral polarity can't have priority.

\item[Irrealis blocking] Neutralizing the non-neutral polarity ($P/N$) by an irrealis blocker $(I)$: 
\begin{equation}
I\ P/N\rightarrow O
\end{equation}

The blocker such as modal ``would'' or connector ``if'' can set up a context about possibility of some polarities not necessarily expressed by the author. As a result, a literal polarity is canceled.
\end{description}

\begin{figure}[t]
    \centering
    \subfigure[Polarity propagation.]{
        \includegraphics[width=0.19\linewidth]{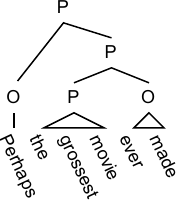}
        \label{fig:polarity_propagation}
    }
    \subfigure[Negation.]{
        \includegraphics[width=0.28\linewidth]{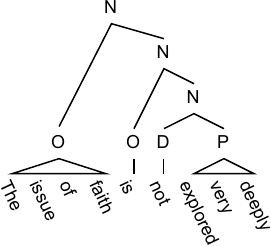}
        \label{fig:negation}
    }
    \subfigure[Conflict resolution.]{
        \includegraphics[width=0.19\linewidth]{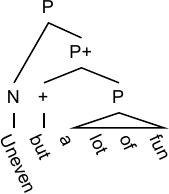}
        \label{fig:conflict_resolution}
    }
    \subfigure[Irrealis blocking.]{
        \includegraphics[width=0.9\linewidth]{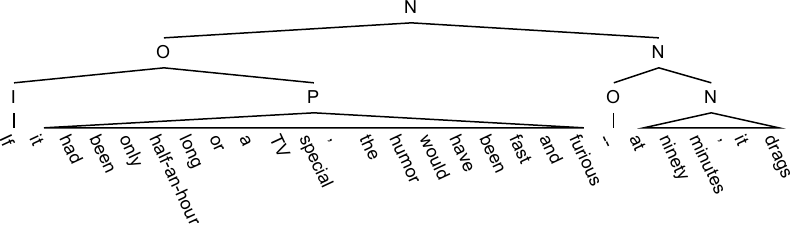}
        \label{fig:irrealis_blocking}
    }
    \caption{Examples of different binary compositions.}
    \label{fig:compositions}
\end{figure}

The full binary rule list is shown in Table \ref{tab:binary-rull-list} in Appendix \ref{app:binary_rules}\footnote{Readers might ask that why explicit triggers are involved in some rules, for example, we can just define a general ``glue'' rule $P\ N\rightarrow P/N$ to handle conflict resolution instead of defining the modifier ($+/-$) to trigger the priority modification, as done by \citet{dong2015statistical}. This is because when only the root label annotation is available, this general rule is easily abused so that the semantic tree degenerates to the sentiment tree as a consequence. The optimal binary rule should satisfy that the output label is uniquely determined given the input ones, requiring us to attribute each label to the specific composition as detailed as possible.}. We also present examples of those compositions Figure \ref{fig:compositions}. Those compositions appears very commonly. To illustrate this, we randomly sample 100 examples in SST2 and MR and count occurrences of above compositions, where 97 and 98 examples in SST2 and MR can be explained by the above compositions. Thus, we believe our rules can cover most cases.

The second type is terminal-unary rule defining the legal preterminals of single words, which is in the form of $\omega\rightarrow A$ ($A\in{\cal N}_{\rm pret}=\{N,P,O,D,I,+,-\},\omega\in{\cal V}$). As introduced, $A$ can't be the polarity priority ($N^+/N^-/P^+/P^-$).

We further define the preterminal-unary rule as the third type, including rules $A\rightarrow A$ ($A\in\{P,N,O,D,I,+,-\}$) and $D/I/+/-\rightarrow O$. Those rules can only and must appear on the second layer of the semantic tree, which is designed to cancel the function of misrecognized function constituents, leading to better performance in our experiments.

\begin{table}[]
\centering
\small
\begin{tabular}{lcc}
\toprule
\multicolumn{1}{c}{Composition} & SST2 & MR \\ \midrule
Polarity propagation            & 97   & 96 \\
Negation                        & 18   & 18 \\
Conflict resolution             & 18   & 20 \\
Irrealis blocking            & 6    & 9  \\
None of the above               & 3    & 4  \\ \bottomrule
\end{tabular}
\caption{The number of existences in the sampled 100 sentences in SST2 and MR.}
\label{tab:composition_num}
\end{table}


\subsection{Sentiment Composition Module}
We now answer the second question: How to model the parser $p(t|x)$ and compute the classifier $p(y|x)$. We show that this process naturally lead to the sentiment composition module.

\subsubsection*{Semantic Tree Parser}
First, we represent the semantic tree $t$ of a sentence $x=(x_0,\cdots,x_{T-1})$ by the set of anchored rules \cite{eisner2016inside} consisting of a rule and its location indices:
\begin{equation}
\begin{aligned}
t=\{(B_{ik}C_{kj}\rightarrow A_{ij})_t|1\leq t\leq T-1\}\\
\cup\{(B_i\rightarrow A_i)_t|1\leq t\leq T\}\\
\cup\{(x_i\rightarrow A_i)_t|1\leq t\leq T\}
\end{aligned}
\end{equation}
where $A_{ij}\ (0\leq i<j< T)$ is an anchored node suggesting a label $A$ covering the constituent ranging from $x_i$ to $x_{j-1}$. $A_i$ is short for $A_{i,i+1}$ which is an unary anchored node covering the word $x_i$. Thus, $B_{ik}C_{kj}\rightarrow A_{ij}$, $B_i\rightarrow A_i$ and $x_i\rightarrow A_i$ represent the binary, preterminal-unary, and terminal-unary anchroed rule, respectively. 

The semantic tree parser $p(t|x)$ is defined by a Gibbs distribution on anchored rules in a tree \cite{finkel2008efficient, durrett2015neural}:
\begin{equation}
p(t|x)=\frac{1}{Z(x)}\prod_{a\in t}\phi(a)=\frac{1}{Z(x)}\exp\left(\sum_{a\in t}s(a)\right)
\end{equation}
where $Z(x)$ is the log-partition function for normalization. $\phi(a)>0$ is the potential function of the anchored rule $a$ defined in the exponential form $\exp(s(a))$, where $s(a)$ is the score to rate how comfortable it is for $a$ to appear in the tree. Scores for different types of anchored rules are defined as the sum of a few subscores rating the comfortableness of corresponding substructures.
\begin{equation}
\begin{aligned}
&s(B_{ik}C_{kj}\rightarrow A_{ij})=\\
&\qquad s_{\rm rule}(BC\rightarrow A)+s_{\rm label}(A,x_{ij})+s_{\rm span}(x_{ij})\\
&s(B_i\rightarrow A_i)=\\
&\qquad s_{\rm rule}(B\rightarrow A)+s_{\rm label}(A,x_i)+s_{\rm span}(x_i)\\
&s(x_i\rightarrow A_i)=s_{\rm rule}(x_i\rightarrow A)
\end{aligned}
\end{equation}
Here the scores of binary and pos-unary rules $s_{\rm rule}(BC\rightarrow A)$ and $s_{\rm rule}(B\rightarrow A)$ are scalar parameters. Other scores are modeled by neural networks:
\begin{equation}
\begin{aligned}
&s_{\rm rule}(x_i\rightarrow A)={\bf w}_{\rm rule}^A\cdot {\bf h}_i^{\leq L}+b_{\rm rule}^A\\
&s_{\rm label}(A,x_{ij})={\bf w}_{\rm label}^A\cdot {\bf h}_{ij}^L+b_{\rm label}^A\\
&s_{\rm span}(x_{ij})={\bf w}_{\rm span}\cdot {\bf h}_{ij}^L+b_{\rm span}
\end{aligned}
\label{eq:scores}
\end{equation}
where $\cdot$ is the vector dot product. ${\bf w}_\cdot^\cdot$ and $b_\cdot^\cdot$ are learning parameters. ${\bf h}_{ij}^l$ is the phrase representation of the constituent $x_{ij}$ in the $l$ layer, which is computed by a text encoder $m$:
\begin{equation}
\begin{aligned}
&{\bf h}_0^0, \cdots, {\bf h}_{T-1}^0,\cdots,{\bf h}_0^{L-1}, \cdots, {\bf h}_{T-1}^{L-1}\\
&\qquad\qquad=m({\bf e}_0,\cdots,{\bf e}_{T-1})\\
&{\bf h}_{ij}^l=\frac{\sum_{t=i}^{j-1}{\bf h}_t^l}{j-i}
\end{aligned}
\end{equation}
where ${\bf e}_i$ is the word embedding of $x_i$. Note that we compute $s_{\rm label}$ and $s_{\rm span}$ using top layer phrase representations, but compute $s_{\rm rule}$ using a lower layer one. This is because the recognition of the preterminal is easier than determining if this label is cancelled. Thus the simple phrase representation ${\bf h}_{ij}^{\leq L}$ is sufficient for the former, while the more ``contextual'' one ${\bf h}_{ij}^L$ is in favor by the latter.



    


\subsubsection*{Inducing the Classifier from the Parser}
As shown in Equation (\ref{eq:formalization}), the classifier is induced by marginalizing over all the semantic trees of the input sentence, which can be efficient done by the inside algorithm. To illustrate this, we first let ${\cal T}_x(A_{ij})$ and ${\cal T}_x(B_{ik}C_{kj}\rightarrow A_{ij})$ be sets of subtrees of sentence $x$ that are covered by the anchored node $A_{ij}$ and rule $B_{ik}C_{kj}\rightarrow A_{ij}$, respectively. The inside algorithm defines the inside term $\alpha_x(A_{ij})=\sum_{t\in{\cal T}_x(A_{ij})}\prod_{a\in t}\phi(a)$, which is the sum of the potentials of subtrees covered by $A_{ij}$. The inside term is computed recursively in a bottom-up manner:
\begin{equation}
\begin{aligned}
\alpha_x(&A_i)=\phi(x_i\rightarrow A_i)\sum_{B\rightarrow A\in{\cal R}}\phi(B_i\rightarrow A_i)\\
\alpha_x(&A_{ij})=\\
&\sum_{BC\rightarrow A\in{\cal R}\atop i<k<j}\phi(B_{ik}C_{kj}\rightarrow A_{ij})\alpha_x({B_{ik}})\alpha_x(C_{kj})
\end{aligned}
\label{eq:inside}
\end{equation}
where $\alpha_x(A_i)$ is the initial value of this recursion. Obvious, the time complexity of the inside algorithm is $O(|{\cal R}|T^3)$. It can be shown that the inside term of the root anchored node $\alpha_x(A_{0T})$, abbreviated as $\alpha_x(A)$, equals to the unnormalized probability that the root of the semantic tree is $y$. Thus, we have:
\begin{equation}
\begin{aligned}
&p(y=A|x)=\frac{\alpha_x(A)}{\sum_{B\in{\cal Y}}\alpha_x(B)}\\
&\qquad =\frac{\exp(s_{\rm label}(A,x)+s_{\rm SCM}(A,x))}{\sum_{B\in{\cal Y}}\exp(s_{\rm label}(B,x)+s_{\rm SCM}(B,x))}\\
&s_{\rm SCM}(A,x)=\mathop{\rm logsumexp}_{BC\rightarrow A\in{\cal R}\atop 0<k<T}\left(s_{\rm rule}(BC\rightarrow A)\right.\\
&\qquad\qquad\qquad\left.+\log\alpha_x(B_{0k})+\log\alpha_x(C_{kT})\right)
\end{aligned}
\end{equation}
As seen, the logit in the softmax includes an extra score $s_{\rm SCM}(A,x)$ as a complement to the regular one $s_{\rm label}(A,x)$, where the former and the latter can be understood as the accordance of assigning the label $A$ by means of sentiment composition and pattern recognition, respectively. Thus, we call $s_{\rm label}$ and $s_{\rm SCM}$ as the recognition module and the sentiment composition module, respectively. While the recognition module is only learned from the data, the sentiment composition module incorporates general and invariant human knowledge in the form of sentiment composition rules, which is more robust for domain adaptation, as we shall see in Section \ref{sec:domain_adapt}. 

The last issue is that the proposed SCM is intractable for long documents due to the cube time complexity over length. So for a document, we first cut it into sentences, and then compute their individual logits. Document logits are aggregated by attention on those sentence logits, where attention weights are computed by sentence representations.



\subsection{Training \& Testing}
Now we've obtained the induced classifier, we can apply supervised training by minimizing:
\begin{equation}
{\cal L}_{\rm cls}=-\frac{1}{N}\sum_{n=1}^N\log p(y^n|x^n)
\end{equation}
This objective might be enough for the classification, but not for a plausible semantic tree explanation. Cases in which a semantic tree can reach a right root label with wrong preterminals and improper structure do exist. For example, if we choose BERT \cite{devlin2018bert} as the encoder, the method might assign non-neutral polarity to \texttt{[CLS]}, and recognize any other tokens as neutral polarity, since \texttt{[CLS]} representation is usually treated as the sentence representation. An effective way to improve the plausibility is to learn the explanation via more explicit annotations \cite{strout2019human, zhong2019fine}, even if those annotations are weak or incomplete. Therefore, we additional introduce two objectives to regularize the tree.

For the preterminal plausibility, we construct a lexicon to annotate the preterminal sequence of each sentence and conduct weakly-supervised learning on the annotation. As introduced, there are 7 preterminals in the proposed grammar, 3 sentimental and 5 functional. We utilize sentiwordnet \cite{baccianella2010sentiwordnet} and stopwords in NLTK\footnote{\url{https://www.nltk.org/}} and spaCy\footnote{\url{https://spacy.io/}} library to annotate non-neutral and neutral sentimental labels, respectively. For functional labels, we manually build a lexicon based on irrealis blockers and priority modifiers from \citet{taboada2011lexicon}, and negators in \citet{loughran2011liability}. The functional lexicon is shown in Table \ref{tab:functional_lexicon} in Appendix \ref{app:functional_lexicon}. Let $o^n$ be the annotated preterminal sequence of the sentence $x^n$, and ${\cal S}^n$ be the set containing the indices of all annotated words. Then, we optimize the following conditional log-likelihood based on the terminal-unary score function in Equation (\ref{eq:scores}):
\begin{equation}
\begin{aligned}
&{\cal L}_{\rm pos}=-\frac{1}{\sum_{n}^N|{\cal S}^n|}\sum_{i\in{\cal S}^n}\log q(o^n_i|x^n)\\
&q(o_i|x)=\frac{\exp(s_{\rm rule}(x_i\rightarrow o_i))}{\sum_{A\in{\cal N}_{\rm pos}} \exp(s_{\rm rule}(x_i\rightarrow A))}
\end{aligned}
\end{equation}

For the structural plausibility, we annotate the syntactical tree for each sentence through Berkeley parser \cite{kitaev2018constituency, kitaev2018multilingual}, which is a SOTA parser based on T5 \cite{raffel2020exploring} and trained on the Penn Treebank (PTB) \cite{taylor2003penn}. We convert the tree to the form of left-branching chomsky normal form (CNF) \cite{chomsky1963formal}, and omit non-terminal labels to obtain the tree skeleton. Our goal is to make the semantic tree structure resemble the annotated PTB tree structure. Given the annotated skeleton $k^n$ of the sentence $x^n$, we minimize the conditional likelihood:
\begin{equation}
\begin{aligned}
&{\cal L}_{\rm str}=-\frac{1}{N}\sum_{n=1}^N\log r(k^n|x^n)\\
&r(k|x)=\frac{1}{Z'(x)}\prod_{c\in k}\exp\left(\sum_{c\in k} s_{\rm span}(c)\right)
\end{aligned}
\end{equation}
where $c$ is a span in the skeleton $k$. As seen, $r(k|x)$ is defined by a Gibbs distribution with span score functions in Equation (\ref{eq:scores}). The normalization term $Z'(x)$ is also computed via the inside algorithm similar to Equation (\ref{eq:inside}).

The final objective is the linear combination of the above three objectives\footnote{Note that both plausibility objectives are conducted on incomplete annotations of the semantic tree. The principled way is to learn the distribution of these annotations conditioned on the input sentence, which is induced from the parser $p(t|x)$ by marginalizing over all the remained unannotated structures. However, this marginalization is intractable in our case. Thus, here we only approximate the true distribution with the product of the expected counts.}:
\begin{equation}
{\cal L}=\omega_{\rm cls}{\cal L}_{\rm cls}+\omega_{\rm pos}{\cal L}_{\rm pos}+\omega_{\rm str}{\cal L}_{\rm str}
\end{equation}

When the model is well-trained, it is able to not only predict the sentiment label but also generate the semantic tree as the explanation:
\begin{equation}
\begin{aligned}
y^\star&=\mathop{\arg\max}_{y\in{\cal Y}}p(y|x)\\
t^\star&=\mathop{\arg\max}_{t\in{\cal T}_x(y^\star)}p(t|x)
\end{aligned}
\end{equation}
The second argmax is to decode the best semantic tree with the maximal conditional probability, which is solved by the CKY algorithm \cite{kasami1965efficient, daniel1967recognition}. 

\section{Experiments}
\label{eq:experiment}
In this section, we conduct experiments to illustrate that the proposed SCM module is able to improve the accuracy performance. 

\begin{table}[t]
\centering
\small
\begin{tabular}{lccc}
\toprule
\multicolumn{1}{c}{\multirow{2}{*}{Method}} & \multirow{2}{*}{MR} & \multicolumn{2}{c}{SST2} \\ \cmidrule{3-4}
\multicolumn{1}{c}{} & & sentence & phrase \\ \midrule
\multicolumn{3}{l}{\textit{\footnotesize Sequential models}}        \\
BiLSTM (\citeyear{hochreiter1997long})                     & 83.27  & 87.52     & 89.68       \\
BERT (\citeyear{devlin2018bert})                       & 87.65  & 92.25     & 93.52       \\ \midrule
\multicolumn{3}{l}{\textit{\footnotesize Sentiment tree models}} \\
MVRNN (\citeyear{socher2013recursive})& -   & - & 82.90      \\
RNTN (\citeyear{socher2013recursive}) & -   & - & 85.40      \\
BiTreeLSTM (\citeyear{teng2017head})  & -   & - & 90.30       \\
RTCM (\citeyear{zhang2019tree})       & -   & - & 90.30     \\
TreeLSTM+WG (\citeyear{zhang2019latent})       & -   & - & 89.70 \\
TreeLSTM+LVG (\citeyear{zhang2019latent})      & -   & - & 89.80  \\
TreeLSTM+LVeG (\citeyear{zhang2019latent})     & -   & - & 89.80  \\ \midrule
\multicolumn{3}{l}{\textit{\footnotesize Untagged tree (by external parser) models}}\\
MVRNN (\citeyear{socher2012semantic}) & 79.00  & -     & -           \\
TreeLSTM (\citeyear{tai2015improved}) & 78.70  & 88.00     & -           \\
\cite{liu2017adaptive}     & 81.90      & 87.80     & -        \\
\cite{liu2017dynamic}      & 81.70      & 87.80     & -    \\
\cite{kim2019dynamic}      & 83.80   & -         & 91.30       \\ \midrule
\multicolumn{3}{l}{\textit{\footnotesize Latent untagged tree models}}       \\
RL-SPINN (\citeyear{yogatama2016learning}) & -       & -         & 86.50       \\
Gumbel-Tree (\citeyear{choi2018learning}) & -       & -         & 90.70       \\
\cite{havrylov2019cooperative}     & -       & -         & 90.20       \\
CRvNN (\citeyear{chowdhury2021modeling})   & -       & -         & 88.30       \\ \midrule
\multicolumn{3}{l}{\textit{\footnotesize Latent semantic tree models (Ours)}}\\
BiLSTM+SCM                 & 83.41   & 88.03     & 90.06       \\
BERT+SCM                  & \textbf{88.16}     & \textbf{92.31}     & \textbf{93.96}       \\ \bottomrule
\end{tabular}
\caption{Sentiment classification accuracy results.}
\label{tab:sentiment-classification}
\end{table}

\subsection{Datasets}
We adopt MR \cite{pang2005seeing} and SST2 \cite{socher2013recursive} in this experiment. MR contains 10662 movie reviews, half of which are positive/negative. Since it has no train/dev/test splits, we follow the convenience to conduct 10-fold cross validation. SST2 is built from SST by binarizing the 5-class sentiment label. Common settings of SST2 include SST2-S which only uses the sentence for training, and SST2-P which uses all labeled non-neutral phrases for training, of which the training size is 6920 and 98794, respectively. In both settings, there are 872/1821 sentences for validation/testing. 

\subsection{Implementation}
We utilize BiLSTM \cite{hochreiter1997long} and BERT \cite{devlin2018bert} (base version) as backbone encoders for modeling the constituent representations. For both models, we use the first layer representations to compute the terminal-unary scores. We use momentum-based gradient descent \cite{qian1999momentum} (we set the momentum to be 0.9), along with cosine annealing learning rate schedule \cite{loshchilov2016sgdr} to optimize our models. For detailed hyper-parameter settings, please check the configuration files in our publicly available repository.

\subsection{Baselines}
Compared models include sequential models and three types of tree models: sentiment tree models, untagged tree models and latent untagged tree models. Both tree models ultilize recursive neural networks (RvNNs) \cite{socher2011parsing} for modeling phrases in the sentence following a tree structure. Sentiment tree models have the full sentiment tree supervision, and learned to predict labels of all nodes in the tree. By contrasts, tree structures for untagged tree models are obtained by an external parser, and only the root node label is available for training. Latent untagged tree models learn to generate the tree structure itself, which is implicit supervised by the task objectives.

\subsection{Results}
We report the accuracy of different models in Table \ref{tab:sentiment-classification}, which we can find that: 1) Compared to the original sequential model, we can see that adding the proposed SCM steadily improves the classification accuracy for both BiLSTM and BERT encoder all the datsets and settings, directly reflecting the effectiveness of our method. 2) Armed with the proposed SCM, the sequential BiLSTM achieves better or competitive performance with previous tree models on both datasets and settings. Specially, it outperforms each baselines on SST-2. This might suggest that the hierarchical RvNN is not necessarily the best way to model compositions, which a flat sequential model could do just as well. 3) We also admit that the performance improvement from our method is not that huge, which our BiLSTM model doesn't surpass all compared models on MR and SST2-P. However, since our motivation is interpretability, we believe that the performance is sufficient.

\section{Discussion}

\subsection{Sentiment Domain Adaptation}
We conduct experiments in the cross-domain setting. \label{sec:domain_adapt}
We adopt Amazon in this experiment. Amazon is a widely-used domain adaption dataset collected by \citet{blitzer2007biographies}. It contains review documents from the Amazon website in four domains: Books (B), DvDs (D), Electronics (E) and Kitchen \& Housewares (K), where each domain contains 2000 labeled reviews. Following previous works, the model is trained on one domain and tested on the other three domains, yielding 12 cross-domain sentiment classification subtasks. For each subtask, we randomly sample 1600 examples in the source domain for training, and left the other 400 examples for validation.

\begin{table}[t]
\centering
\small
\begin{tabular}{c|p{1cm}<{\centering}p{1cm}<{\centering}|p{1cm}<{\centering}p{1cm}<{\centering}}
\toprule
\multirow{2}{*}{S$\rightarrow$T} & \multirow{2}{*}{BiLSTM} & BiLSTM & \multirow{2}{*}{BERT} & BERT \\ 
                 &                & +SCM                    &                & +SCM                 \\ \midrule
B$\rightarrow$D & 82.65           & \textbf{82.75}         & 88.96          & \textbf{89.95}       \\
B$\rightarrow$E & 76.50           & \textbf{79.60}         & 86.15          & \textbf{87.70}       \\
B$\rightarrow$K & \textbf{78.05}  & 77.75                  & \textbf{89.05} & 87.65       \\
D$\rightarrow$B & 80.80           & \textbf{82.35}         & \textbf{89.40} & 88.05       \\
D$\rightarrow$E & 77.05           & \textbf{80.85}         & 86.55          & \textbf{87.55}       \\
D$\rightarrow$K & 77.65           & \textbf{79.85}         & 87.53          & \textbf{88.30}       \\
E$\rightarrow$B & 73.85           & \textbf{75.45}         & 86.50          & \textbf{86.75}       \\
E$\rightarrow$D & 77.25           & \textbf{78.25}         & \textbf{87.95} & 87.30       \\
E$\rightarrow$K & \textbf{84.85}  & 83.90                  & 91.60          & \textbf{91.85}       \\
K$\rightarrow$B & 71.65           & \textbf{75.80}         & \textbf{87.55} & 86.35       \\
K$\rightarrow$D & 73.75           & \textbf{76.50}         & \textbf{87.30} & 87.25       \\
K$\rightarrow$E & \textbf{82.95}  & 82.90                  & 90.45          & \textbf{90.80}       \\ \midrule
Average         & 78.08           & \textbf{79.66}         & 88.25          & \textbf{88.29}       \\ \bottomrule
\end{tabular}
\caption{Domain adaptation results on Amazon.}
\label{tab:domain-adaptation}
\end{table}
We report the accuracy of different subtask in Table \ref{tab:domain-adaptation}. As seen, compared to original sequential models, adding the proposed SCM improves the adaptation accuracy in most cases and on average as well, especially for BiLSTM which is trained from scratch. The improvement originates from the injected domain-invariant human knowledge in the proposed SCM, which helps the model to be less sensitive to the domain. The performance improvement of pretrained model BERT is not that significant because the pretraning process has already given the generalization ability to it.

\begin{table}[t]
\centering
\small
\begin{tabular}{lcccc}
\toprule
\multicolumn{1}{c}{Method}                        & Acc   & Tree F1 \\ \midrule
CNF                                               & -     & 77.19  \\ \midrule
BiLSTM                                            & 87.52 & -      \\
$+{\cal G}$                                       & 87.10 & 21.38  \\
$+{\cal G}+{\cal L}_{\rm pos}$                    & 87.59 & 21.04  \\
$+{\cal G}+{\cal L}_{\rm str}$                    & 86.77 & \textbf{55.04}  \\
$+{\cal G}+{\cal L}_{\rm pos}+{\cal L}_{\rm str}$ & \textbf{88.03}  & 46.85 \\ \midrule
BERT                                              & 92.25 & -       \\
$+{\cal G}$                                       & 91.32 & 09.56  \\
$+{\cal G}+{\cal L}_{\rm pos}$                    & 91.82 & 12.28  \\
$+{\cal G}+{\cal L}_{\rm str}$                    & 91.93 & \textbf{51.05} \\
$+{\cal G}+{\cal L}_{\rm pos}+{\cal L}_{\rm str}$ & \textbf{92.31} & 50.94 \\ \bottomrule
\end{tabular}
\caption{Ablation study results on SST2-S. CNF represents the CNF equivalent of the constituency tree generated by Berkeley parser. Its tree F1 is the upper limit of this value.}
\label{tab:ablation-study}
\end{table}

\subsection{Ablation Study}
We conduct ablation study on SST2-S to study effects of different components including the grammar and two plausibility objectives. We report the accuracy and the unlabeled tree F1 of the generated semantic tree w.r.t. PTB trees generated by Berkeley parser for each model in Table \ref{tab:ablation-study}.

We find that the grammar doesn't work out alone when two plausibility objectives are absent, where the accuracy drops compared to the original encoder. We speculate this is due to lack of direct information of function labels, making it easier to mis-recognition on those labels. Such error would accumulated from bottom to up in the tree and pollute other sentences including the same constituent, causing the performance drop.

\begin{figure}[t]
\centering
\includegraphics[width=0.5\textwidth]{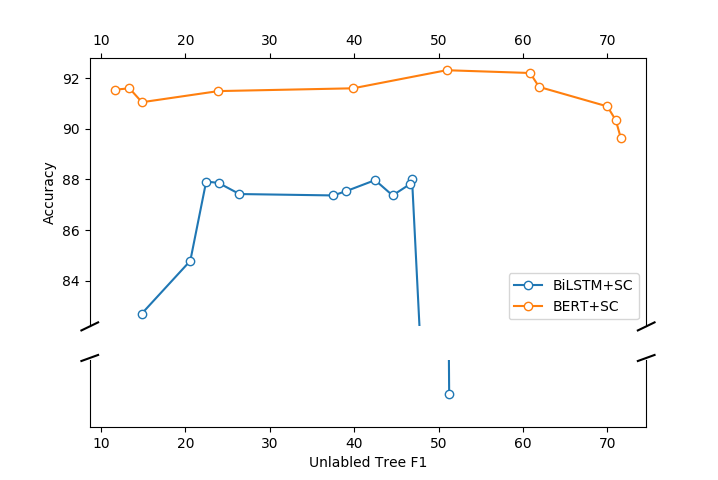}
\caption{The relation between the accuracy and unlabeled tree F1 on SST2-S.}
\label{fig:f1-acc}
\end{figure}

The preterminal plausibility objective ${\cal L}_{\rm pos}$ alleviates this issue effectively with an obvious performance improvement for both encoders. For the structure plausibility objective ${\cal L}_{\rm str}$, though it makes the tree structure more syntactically meaningful with higher unlabeled tree F1, it doesn't necessarily guarantee the performance improvement. This suggests that the optimal tree structure might not exactly resemble PTB tree structure. On the contrary, the tree structure learned without ${\cal L}_{\rm str}$, which has little similarity with PTB tree structure, is also suboptimal with mediocre accuracy. To study the optimal tree structure, we alter the balancing factor $\omega_{\rm str}$ and obtain models with different unlabeled tree F1 w.r.t. PTB trees and accuracy. Then, we visualize relation between these two metrices in Figure \ref{fig:f1-acc}. We can see that accuracy roughly shows a trend of first increasing and then decreasing when the tree gets more syntactical meaningful for both encoders (i.e., has higher unlabeled tree F1). This is contrary to that of \citet{williams2018latent} which finds that the optimal tree structure of untagged tree methods RL-SPINN \cite{yogatama2016learning} and Gumbel-Tree \cite{choi2018learning} do not resemble PTB tree structure. This might because our method has a specific grammar with syntactical information restraining the tree structure, while untagged tree methods accommodate for any structure.



\begin{table}[t]
\centering
\small
\begin{tabular}{lcc}
\toprule
Method                      & \multicolumn{1}{c}{Grammar} & Acc   \\ \midrule
\multirow{2}{*}{BiLSTM+SCM} & glue    & 87.42 \\
                            & SCG     & \textbf{88.03} \\ \midrule
\multirow{2}{*}{BERT+SCM}   & glue    & 92.20 \\
                            & SCG     & \textbf{92.31} \\ \bottomrule
\end{tabular}
\caption{Accuracy performances of different grammars on SST2-S.}
\label{tab:SCG_effects}
\end{table}

\subsection{Effects of SCG}
To show the effectiveness of the proposed SCG, we compare it with the glue grammar \cite{taboada2011lexicon} whose binary rules are very free and in the form ${BC\rightarrow A}$ ($A,B,C\in\{P,N,O\}$). Such rules act like the glue to connect adjacent constituents with any polarities. The results are shown in Table \ref{tab:SCG_effects}, which our proposed SCG is more effective with better accuracy compared to the glue grammar. We think this is because glue grammar rules are too free to carry specific sentiment composition knowledge, which is is helpless for the task.

\begin{figure}[t]
    \centering
    \subfigure[Double irealistic blocking.]{
        \includegraphics[width=0.64\linewidth]{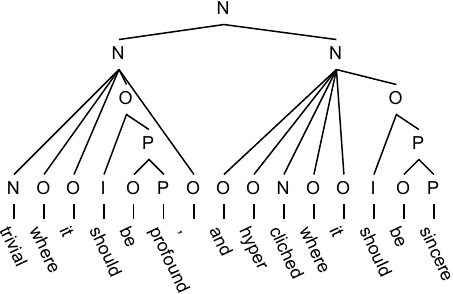}
        \label{fig:double_irrealis_blocking}
    }
    \subfigure[Negation in conflict resolution.]{
        \includegraphics[width=0.21\linewidth]{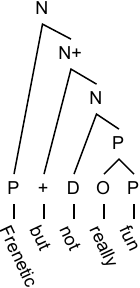}
        \label{fig:negation}
    }
    \caption{Semantic trees of compound sentiment compositions, generated by BiLSTM+SCM. We flatten the polarity propagation rules for compactness.}
    \label{fig:compound_structure}
\end{figure}

\subsection{Qualitative Study}
We qualitatively show a few examples to show our method can handle compound sentiment compositions in Figure \ref{fig:compound_structure}. The first case is a sentence with two negative constituents joining by a coordinating conjunction, each of which has an irrealis blocking within. The second case is a sentence with negation under conflict resolution. For both cases, the prediction is not simple since the model is susceptible to the surface and literal meaning in the sentence, which might interfere the correct decision. Taking the sentiment composition explicitly, we can see that our method successfully judge the semantic role of different constituents, and finally compose plausible tree explanations.

\section{Related Works}
Sentiment composition is one of the key to sentiment analysis, which considers the semantic of a constituent from both recognition and composition views \cite{polanyi2006contextual, moilanen2007sentiment}. That is, it decomposes the classification of a sentence into a hierarchical tree structure explicitly showing how the polarity of the sentence come from the composition of its sub-constituents. Early works are mainly based on manual rules and semantic lexicon that is constructed either manually \cite{wilson2005recognizing, kennedy2006sentiment} or automatically \cite{dong2015statistical, toledo2018learning}. Nowadays, represented via different forms of tree, sentiment composition is often learned explicitly or implicitly in the end-to-end learning manner of neural network models. 

Common tree forms include untagged tree and sentiment tree, while the learning paradigm is also varied in literature. To be concrete, untagged tree can either be directly obtained from the external syntactic parser \cite{socher2012semantic, tai2015improved, liu2017adaptive, liu2017dynamic, kim2019dynamic}, or serve as a latent variable learned implicitly \cite{yogatama2016learning, maillard2018latent, choi2018learning, havrylov2019cooperative, chowdhury2021modeling}. Compared to the untagged one, sentiment tree offers more information about sentiment polarity of each constituent in the tree. As the most representative resource in this form, SST \cite{socher2013recursive} formalizes sentiment composition as a parsing task, motivating lots of works to learn the tree supervisedly \cite{teng2017head, zhang2019tree, zhang2019latent}. Sentiment tree is also a popular explanation form for post-hoc interprebility since it can provide hierahical attribution scores \cite{chen2020generating, zhang2020interpreting}. While both existing forms are useful, they are suboptimal due to their in-ability to explicitly interpret sentiment composition, which our proposed semantic tree fills this gap.

\section{Conclusions}
In this paper, we present semantic tree to explicitly interpret sentiment compositions in sentiment classification. we carefully design a grammar under each compositions from the linguistic inspiration, and learn to extract semantic tree explanations without full annotations. Quantitative and qualitative results demonstrate that our method is effective and can generate plausible tree explanations.

\section{Limitations \& Ethics Statement}
Our method is first limited by the proposed grammar that doesn't cover all the realistic cases. As shown in Table \ref{tab:composition_num}, there are still a few cases in the randomly sampled 100 examples that none of the defined rules can explain. Secondly, the time complexity of our method is the cube of the sentence length, limiting its direct applications on long documents. So we have to classify the document based on classification of individual sentences, which might be problematic since the sentiment of different sentences in the document may affect each other.

All the experiments in this paper are conducted on public available datasets, which has no data privacy concerns. Meanwhile, this paper doesn't involve human annotations, so there are no related ethical concerns.

\section*{Acknowledgements}
This work was supported by the National Key R\&D Program of China (2022ZD0160503) and the National Natural Science Foundation of China (No.61976211, No.62276264), and the Strategic Priority Research Program of Chinese Academy of Sciences (No.XDA27020100). This research was also supported by Meituan.

\bibliography{anthology,custom}
\bibliographystyle{acl_natbib}

\appendix

\section{Binary Rules}
\label{app:binary_rules}

Table \ref{tab:binary-rull-list} shows all the binary rules contained in the proposed SCG.

\section{Functional Lexicon}
\label{app:functional_lexicon}
Table \ref{tab:functional_lexicon} lists functional lexicon in the manually constructed lexicon.

\begin{table}[]
\small
\centering
\begin{tabular}{lll}
\toprule
\multicolumn{1}{c}{Composition} &
  \multicolumn{2}{c}{Rules} \\ \midrule
Polarity propagation &
  \begin{tabular}[c]{@{}l@{}}$O\ P\rightarrow P$\\ $P\ O\rightarrow P$\\ $O\ N\rightarrow N$\\ $N\ O\rightarrow N$\end{tabular} &
  \begin{tabular}[c]{@{}l@{}}$O\ O\rightarrow O$\\ $P\ P\rightarrow P$\\ $N\ N\rightarrow N$\end{tabular} \\ \midrule
Negation &
  \begin{tabular}[c]{@{}l@{}}$D\ P\rightarrow N$\\ $D\ N\rightarrow P$\end{tabular} &
  \begin{tabular}[c]{@{}l@{}}$P\ D\rightarrow N$\\ $N\ D\rightarrow P$\end{tabular} \\ \midrule
Conflict resolution &
  \begin{tabular}[c]{@{}l@{}}$P\ +\rightarrow P^+$\\ $N\ +\rightarrow N^+$\\ $+\ P\rightarrow P^+$\\ $+\ N\rightarrow N^+$\\ $P^+\ +\rightarrow P^+$\\ $N^+\ +\rightarrow N^+$\\ $+\ P^+\rightarrow P^+$\\ $+\ N^+\rightarrow N^+$\\ $P\ -\rightarrow P^-$\\ $N\ +\rightarrow N^-$\\ $-\ P\rightarrow P^-$\\ $-\ N\rightarrow N^-$\\ $P^-\ -\rightarrow P^-$\\ $N^-\ +\rightarrow N^-$\\ $-\ P^-\rightarrow P^-$\\ $-\ N^-\rightarrow N^-$\\ $P^+\ O\rightarrow P^+$\\ $N^+\ O\rightarrow N^+$\\ $P^-\ O\rightarrow P^-$\\ $N^-\ O\rightarrow N^-$\\ $O\ P^+\rightarrow P^+$\\ $O\ N^+\rightarrow N^+$\\ $O\ P^-\rightarrow P^-$\\ $O\ P-\rightarrow P^-$\end{tabular} &
  \begin{tabular}[c]{@{}l@{}}$N\ P^+\rightarrow P$\\ $N^-\ P^+\rightarrow P$\\ $N^-\ P\rightarrow P$\\ $P^+\ N\rightarrow P$\\ $P^+\ N^-\rightarrow P$\\ $P\ N^-\rightarrow P$\\ $N\ P^-\rightarrow N$\\ $N^+\ P^-\rightarrow N$\\ $N^+\ P\rightarrow N$\\ $P^-\ N\rightarrow N$\\ $P^-\ N^+\rightarrow N$\\ $P\ N^+\rightarrow N$\end{tabular} \\ \midrule
Irrealis blocking &
  \begin{tabular}[c]{@{}l@{}}$I\ P\rightarrow O$\\ $I\ N\rightarrow O$\end{tabular} &
  \begin{tabular}[c]{@{}l@{}}$P\ I\rightarrow O$\\ $N\ I\rightarrow O$\end{tabular} \\ \bottomrule
\end{tabular}
\caption{Binary rules in the proposed SCG.}
\label{tab:binary-rull-list}
\end{table}

\begin{table}[]
\small
\centering
\begin{tabular}{lp{4.5cm}}
\toprule
\multicolumn{1}{c}{Label} & \multicolumn{1}{c}{Words}                                        \\ \midrule
Priority riser $+$        & but, however, yet, whereas, still                                \\ \midrule
Priority reducer $-$      & although, though, despite, regardless, nevertheless, nonetheless \\ \midrule
Irrealis blocker $I$    & could, should, would, ought, supposed, if                        \\ \midrule
Negator $D$ &
  no, not, n't, neither, nor, never, none, lack, without, cannot, aint, arent, barely, cant, couldnt, didnt, doesnt, dont, hardly, havent, few, isnt, merely, never, nothing, nobody, shouldnt, wasnt, werent, wont, wouldnt \\ \midrule
\end{tabular}
\caption{Funtional lexicon.}
\label{tab:functional_lexicon}
\end{table}

\end{document}